\documentclass[]{article}
\usepackage[margin=2.8cm]{geometry}
\usepackage{graphicx}
\usepackage{kpfonts}
\usepackage{natbib}
\usepackage{hyperref}
\usepackage{cleveref}

\usepackage[]{authblk}
\usepackage{fancyhdr}
\fancypagestyle{firstpage}{\fancyhead[L]{}\fancyhead[R]{}}

\setcitestyle{square,numbers}
\newcommand{\furl}[1]{\footnote{\url{http://#1}}}

\title{Rethinking Large Language Models \\in Mental Health Applications}

\author[$\dag$]{Shaoxiong Ji }
\author[$\star$]{Tianlin Zhang }
\author[$\star$]{Kailai Yang }
\author[$\star$]{Sophia Ananiadou }
\author[$\alpha$]{Erik Cambria }

\affil[$\dag$]{University of Helsinki, Finland}
\affil[$\star$]{The University of Manchester, UK}
\affil[$\alpha$]{Nanyang Technological University, Singapore}
\affil[ ]{Email:~shaoxiong.ji@helsinki.fi;~\{kailai.yang,tianlin.zhang\}@postgrad.manchester.ac.uk;\\sophia.ananiadou@manchester.ac.uk, cambria@ntu.edu.sg}

\date{}

\begin{document}

\maketitle
\thispagestyle{firstpage}

\begin{abstract}
Large Language Models (LLMs) have become valuable assets in mental health, showing promise in both classification tasks and counseling applications. 
This paper offers a perspective on using LLMs in mental health applications. 
It discusses the instability of generative models for prediction and the potential for generating hallucinatory outputs, underscoring the need for ongoing audits and evaluations to maintain their reliability and dependability.
The paper also distinguishes between the often interchangeable terms ``explainability'' and ``interpretability'', advocating for developing inherently interpretable methods instead of relying on potentially hallucinated self-explanations generated by LLMs.
Despite the advancements in LLMs, human counselors' empathetic understanding, nuanced interpretation, and contextual awareness remain irreplaceable in the sensitive and complex realm of mental health counseling. 
The use of LLMs should be approached with a judicious and considerate mindset, viewing them as tools that complement human expertise rather than seeking to replace it.
\end{abstract}

\section{Introduction}
\label{sec:introduction}

\begin{figure}
    \centering
    \includegraphics[width=0.8\textwidth]{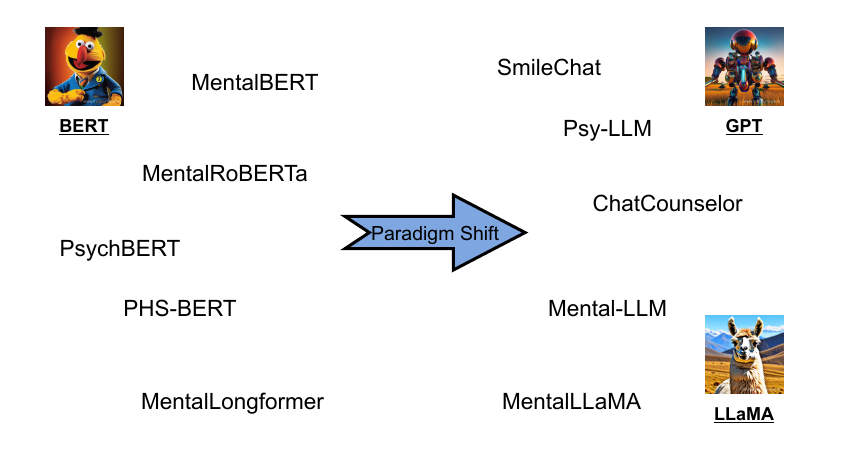}
    \caption{A paradigm shift in NLP for mental health applications from masked language models such as BERT to generative language models such as GPT and LLaMA. Images of BERT, GPT, LLaMA are generated by \href{https://www.midjourneyai.ai}{Midjourney AI Art Generator}.}
    \label{fig:paradigm-shift}
\end{figure}

Mental health is important and has been studied by natural language processing (NLP) using text (e.g., social posts and doctor-patient conversations) as the data sources, leading to the development of automatic methods for various applications, including early detection of mental disorders~\citep{zhang2022natural} and mental health counseling~\citep{althoff2016large}. 
Researchers have employed techniques, ranging from traditional feature engineering to automatic feature learning, such as convolutional neural networks, recurrent neural networks, and transformer networks, for mental illness detection and classification~\citep{zhang2022natural}. 
Recent advances utilize pretrained language models (PLMs).
PLMs trained with the masked language modeling objective have become popular for training classification models in this domain. 
Domain-specific continual pre-training has also undergone intensive development to acquire domain knowledge with representative discriminative models including PsychBERT~\citep{vajre2021psychbert}, MentalBERT~\citep{ji2022mentalbert}, PHS-BERT~\citep{naseem-etal-2022-benchmarking}, and MentalLongformer~\citep{ji-domain-specific-longformer}.
A recent shift as in \Cref{fig:paradigm-shift} has occurred towards prompt learning, where generative large language models (LLMs) such as SmileChat~\citep{qiu2023smile}, Psy-LLM~\citep{lai2023psyllm}, Mental-LLM~\citep{xu2023mental}, MentalLLaMA~\citep{yang2023mentalllama}, ChatCounselor~\citep{liu2023chatcounselor}, and MindWatch~\citep{bhaumik2023mindwatch}, are used to generate predictions or counseling based on input prompts related to mental health conditions. 
This shift signifies a growing interest in leveraging generative LLMs and prompt learning for mental health-related tasks.
However, one question looms large: is this a mere hype? This paper delves into the recent developments and concerns surrounding the use of LLMs for early prediction of mental health conditions, generating explanations for mental health conditions, and generating responses in mental health counseling.

The landscape of large language models has undergone substantial transformation in recent years. 
Current LLMs boast hundreds of billions of parameters, a stark contrast to the relatively modest sizes seen in the early 2010s, typically ranging from millions to tens of millions of parameters. 
Notably, models such as BERT, with 110 million parameters, and GPT-2, with 1.5 billion parameters, which were once considered large, now fall into the category of medium-sized language models by standards at the time of writing. 
It is important to note that the size of language models is not the only factor determining their performance. 
Other factors, such as model architecture, training data, and fine-tuning, also play significant roles in their capabilities.
This growth in model size reflects the ongoing evolution of AI language models.
This paper focuses on the recent use of generative LLMs in mental health applications. 
For the purpose of this paper, the term ``LLMs'' refers to generative models trained with the causal language modeling objective, often called next-word prediction in a simpler term.

Our paper offers perspectives on rethinking large language models in the context of mental healthcare. 
When using generation to predict mental health conditions based on a prompt and post, it is worth noting that the generation-as-prediction process can exhibit instability and unpredictability, even with minor changes to the input prompt. 
We discuss empirical results related to this instability and explore theoretical studies on meta-optimization that may underlie this unpredictability. 
Consequently, we advocate carefully auditing generative LLMs when they are used to predict mental health conditions.

When employing LLMs for mental health prediction, a significant concern revolves around the interpretability of the generated output or the so-called explanations. 
LLMs often operate as black-box neural networks, making it challenging to discern how they arrive at their conclusions. 
Therefore, it is essential to emphasize that claims of interpretable mental health analysis should not be taken at face value but substantiated with rigorous proof and verification.
One fundamental concern when using LLMs for mental health prediction is that the generated explanations may not necessarily imply true interpretability.

In the context of early prediction of mental disorders, providing explanations for mental health conditions, and counseling for mental health-related queries, LLMs have the potential to produce incorrect information akin to hallucinations. 
This risk underscores the necessity for further research to assess the reliability and accuracy of these models. 
Developing safeguards and validation mechanisms is essential to minimize the potential for misinformation in mental health applications.
Notably, some LLMs, like LLaMA and BLOOM, have explicitly stated that their use in high-stakes settings is either out of scope or prohibited. 
This underscores the recognition within the AI community of the ethical and practical concerns surrounding the application of LLMs in situations where human well-being and mental health are at stake. 

In conclusion, these limitations and guidelines should serve as a reminder that LLMs are not universally suitable for all mental health applications and should be used judiciously with a full understanding of their strengths and limitations.

\section{Early Prediction Through LLMs' Generation}

Social media platforms have become a rich data source for studying and potentially detecting mental health issues~\citep{pavalanathan2015identity,harrigian2021state,zhang2022natural}. 
Early detection of mental health concerns on social media involves using models to identify signs and patterns that may indicate emotional distress, mental health issues, or potential crises. 
The emergent prompt-based learning follows the steps of pretraining, prompting, and predicting.
For example, an LLM, such as GPT-3~\citep{brown2020language} and its successors, generates the predicted mental disorder label given a prompt and a social post as the input.
The generation-as-prediction paradigm has many advantages in many NLP tasks.
In the mental health analysis, an early evaluation on ChatGPT~\citep{amin2023will} and other LLMs~\citep{yang2023interpretable} indicate LLMs are good generalist models but not as good as specialized discriminative classification models trained specifically for downstream tasks.
Recently, Mental-LLM~\citep{xu2023mental} and MentalLLaMA~\citep{yang2023mentalllama} show that instructional fine-tuning can improve the prediction performance.
However, finetuning generative large models with billions of model parameters still did not outperform discriminative models with millions of parameters.

\paragraph{Instablity of Generation-as-Prediction}

The dynamic nature of generative models means that small alterations in the input prompt can lead to significantly different outputs. 
In the context of mental health prediction, this unpredictability poses a serious concern. 
A minor modification in the wording or framing of a prompt could yield varying and potentially incorrect assessments of an individual's mental health condition. 
For example, \citet{yang2023interpretable} reported that the model's performance is highly sensitive to variations in adjectives describing condition severity while mentioning that few-shot learning could be a possible way to mitigate it. 
Specifically, altering the adjectives of severity from \textit{any}, \textit{some} to \textit{very severe} can result in fluctuations in predictive accuracy without a discernible pattern.
This instability underscores the necessity of thorough audits of the model's performance and response to different inputs.

\paragraph{From the View of Meta Optimization} 

\begin{figure}
    \centering
    \includegraphics[width=0.6\textwidth]{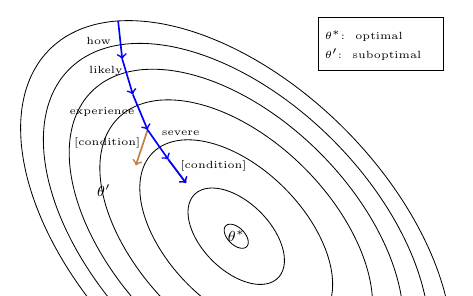}
    \caption{An illustration of prompting from the view of meta update. The change in the prompt might lead to suboptimal, possibly explaining the unpredictable LLMs' generation-as-prediction.}
    \label{fig:meta-opt}
\end{figure}

Some recent research in machine learning has provided insights into the in-context learning behavior of LLMs.
For example, \citet{dai2023can} suggested that LLMs perform implicit gradient descent at inference time.
There are some similar views, such as the concept of learning in-context through gradient descent~\citep{von2023transformers} and the mechanism of causal language modeling through meta-learning~\citep{wu2023meta}.
Meta optimization seems a quite reasonable explanation for the ``learning'' process of LLMs' generation given a prompt.
However, there is no definitive consensus on this matter.
For example, \citet{min-etal-2022-rethinking} showed that ground truth demonstrations are not required for in-context learning, raising the question of whether LLMs might rely on a form of hard memorization.
The debate continues, and a conclusive answer remains elusive.
In the context of unpredictable prediction of LLMs' generation, the optimization process, when viewed as a form of meta-optimization, can appear arbitrary without a certain optimization objective, especially when prompted with free-form inputs as illustrated in \Cref{fig:meta-opt}.
In the case study conducted by \citet{yang2023interpretable}, the adjectives of severity affect the inference time optimization.
This underscores the challenges in adapting LLMs to complex human mental states and the nuances present in self-reported mental health posts.
Overall, the nature of LLMs' in-context learning and the design of prompts remain subjects of ongoing research and debate, given the unique characteristics and challenges posed by LLMs in their generation of text.

\paragraph{Reliablity and Auditing}

Generative language models provide a more flexible and accessible way through API than the preceding pre-training and fine-tuning paradigm, largely due to their rapid development.
OpenAI's ChatGPT, for example, stands as a prominent model accessible via an API, facilitating the creation of generation-as-prediction pipelines for a wide user base.
At the time of writing, despite their larger model size, LLMs utilized for generation-as-prediction still exhibit lower predictive performance than previous models trained with task-specific classification heads~\citep{xu2023mental,yang2023mentalllama}.
Additional (instruction) fine-tuning mitigates this performance gap. 
The significance of fine-tuning on diverse and representative datasets cannot be overstated. 
Addressing biases in the training data and optimizing model hyperparameters to achieve improved performance in mental health classification tasks remain less explored, primarily due to the extensive computing requirements for training large models and searching hyperparameters. 
These factors collectively contribute to the responsible and effective utilization of LLMs in mental health assessment.
Furthermore, the instability of LLMs' generation-as-prediction remains a challenging problem. 
The viewpoint of meta-optimization can probably potentially shed light on the early prediction of mental disorder through LLMs' generation, for example, quantitatively evaluating the equivalent parameter contributions~\citep{lan2019lca} of prompts tailored for mental health applications.
Besides, it is crucial to establish auditing processes~\citep{mokander2023auditing} that assess the model's reliability, sensitivity to input variations, and potential biases to ensure the responsible and accurate use of generative models in mental health applications.
Such audits can help identify and mitigate issues related to unpredictability and instability, ultimately improving the model's suitability for assisting in mental health prediction and ensuring that its outputs are consistent and dependable.

\section{LLM-generated Explanation $\neq$ Interpretablity}
\label{sec:explanations}

While deep learning models are often considered opaque, recent research has unveiled that these hidden representations can, to some extent, offer explanations. 
For instance, there has been ongoing discussion regarding whether attention mechanisms serve as explanations \citep{bibal-etal-2022-attention}, and we acknowledge that there is no definitive consensus on this matter. 
In the context of mental health applications, our stance aligns with the perspective put forth in these publications regarding explainability.
LLMs have the ability of self-explanation to provide explanations for their responses or generate text that clarifies the reasoning behind their answers, which is a form of step-by-step reasoning as referred to chain-of-thought~\citep{wei2022chain}.
However, such explanations can be unfaithful~\citep{xu2023mental} and require targeted efforts for improvements~\citep{turpin2023language}.
Assessing the robustness and faithfulness of LLM-generated explanations in the context of mental health is crucial. 
LLMs may sometimes produce explanations that are overly simplistic or misleading, potentially impacting the quality of mental health interventions. 
It is essential to rigorously evaluate the explanations generated by LLMs to ensure they align with established clinical knowledge and guidelines.
Besides, it is crucial to exercise caution and prudence when making claims about the explainability and interpretability of LLMs-based methods applied to mental health applications.
LLMs' generated explanations do not imply LLMs for mental health analysis are inherently interpretable.
Research works must refrain from using ``interpretability'' and ``explainability'' interchangeably to avoid misconceptions and ensure clarity in discussions surrounding LLMs in mental health applications.

\paragraph{Interpretability and Explanability}

Interpretability and/or explainability are frequently employed in many mental health publications~\citep{joyce2023explainable}. 
 It is crucial to recognize the distinct contrast between ``interpretability'', which pertains to a model's inherent characteristics, and ``explainability'', which refers to the methods used to explain a model or make a model interpretable, while ``explanations'' encompass the actual insights or justifications provided by the model to facilitate users' comprehension of its predictions~\citep{rudin2019stop}.
Despite this, it is worth noting that some literature within the field of LLM uses interpretability and explainability interchangeably, such as \citet{zhao2023explainability}.
Recent work such as \citet{yang2023interpretable} explores how LLMs generate text to explain the prediction of mental disorders.
It is important to recognize that these explanations may lack interpretability.
In other words, LLMs may provide detailed explanations (putting aside the faithfulness aspect for now), but these explanations may not be straightforward or easily comprehensible to human users who seek to understand why the model generates such textual explanations. 
It is crucial to understand that LLMs' explanations, as post-hoc generated text, do not guarantee that the model will be inherently interpretable.
Users may need to exert additional effort or engage in further processing to make sense (or nonsense) of these explanations and render them readily digestible.

\paragraph{Call for Interpretability}

\begin{figure}
    \centering
    \includegraphics{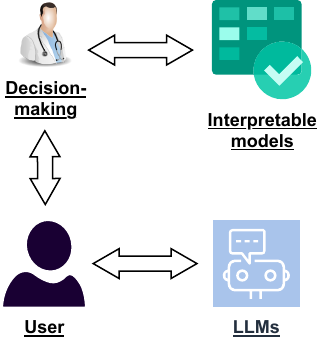}
    \caption{LLMs serve as the user interface to facilitate service quality, while interpretable models are critical for decision-making.}
    \label{fig:LLMs-user-interface}
\end{figure}

LLMs are getting more performant in many applications and improving in mental health applications.
While LLM-based methods are employed for mental health analysis, the claim of interpretability should be considered cautiously.
Our intention is not to dismiss the value of ongoing research on self-explanation. Instead, we aim to clarify definitions and claims, particularly within critical applications like mental healthcare.
One avenue of research in the realm of LLM self-explanation involves engineering techniques or experimental testing that explains the significance of the model's representations and draws intuitive conclusions about the performance of these generated explanations or representations.
In mental health, relying solely on the model-generated explanation is insufficient. 
Human judgment and clinical expertise should be integral in explaining and validating the results.
When explaining the causes behind mental disorders, it is crucial to verify the accuracy and evidence-based nature of the explanations provided by LLMs. 
Additionally, it is essential to carefully monitor and mitigate the potential for LLMs to generate stigmatizing or harmful explanations.
Interpretability is a critical factor, especially in fields where decisions can profoundly affect individuals' well-being~\citep{joyce2023explainable}.
More importantly, we call upon the computational research community in the field of mental health to focus on developing techniques that make these models more inherently interpretable, rigorously define the knowledge being modeled or applied within the mathematical theory, and adhere to proof or analysis that has been done through conceptual representation capacity, generalization, and robustness of neural networks in theory.
The black-box nature of neural networks in LLMs underscores the need for transparency and validation in mental health applications, allowing clinicians and experts to trust and validate the results. 
Although the trade-off between interpretability and accuracy is still a matter of debate, the emphasis on interpretability can help ensure that LLMs become valuable tools in mental health while mitigating the risks associated with their black-box nature.
The future trajectory in integrating LLMs into mental health applications could be ensuring that the LLMs' outputs align with clinical perspectives in interpreting and validating the model's prediction and developing specialized tools tailored for mental health professionals to comprehend the model. 
In this context, data-driven methods like LLMs serve as a user interface to improve the overall usability of the mental health support system and interpretable methods are used for certain aspects of decision-making (\Cref{fig:LLMs-user-interface}). 
An analogy is the well-established diagnostic tools like the nine-item Patient Health Questionnaire (PHQ-9) for assessing depressive symptoms~\citep{kroenke2001phq}. 
Interpretable methods that foster an understanding of their inner workings enable users, especially mental health professionals, to grasp the rationale behind the model's prediction. 

\section{LLMs in Mental Health Counseling}

Chatbots for mental health, such as Woebot, have been developed to provide emotional support and aid in Cognitive Behavioral Therapy (CBT) through conversation with people living in mental health conditions~\citep{fitzpatrick2017delivering}.
\citet{sarkar2023review} reviewed conversational agents for mental health and emphasized clinical knowledge and clinical practice guidelines in making them explainable and safe.
Developments such as MindWatch~\citep{bhaumik2023mindwatch} may play a role in monitoring such risks, but their use should be guided by best practices and ethical considerations.

Here, we discuss whether LLM-generated text is good for counseling in recent studies on empathy, user intention, emotion cause, and beyond. 
We expect reinforcement learning from human feedback to enable helpful and harmless generation, which could possibly enhance LLM-based mental health counseling.
LLM-generated text can have potential applications in counseling, especially in psychological therapies like cognitive behavioral therapy (CBT). 
However, it's essential to approach this with caution. 
The recent development of LLMs might assist in providing information or exercises but should not replace the human element of counseling to offer personalized guidance and adapt interventions to the individual's unique needs, especially in sensitive mental health contexts.

\paragraph{Human Intent, Touch and Empathy}

While LLMs can provide automated responses and information and process long context~\citep{liu2023lost}, the generation mainly relies on learned model parameters from the pretraining corpus and the calculation of the likelihood of the next word. 
They can be distracted by irrelevant context~\citep{shi2023large} and may not fully understand the nuances of individual experiences, especially when there is insufficient individual training data, making their advice less tailored.
These models lack the empathetic and contextual understanding that human counselors possess, which are crucial in counseling, especially in mental health contexts~\citep{sharma2021towards}.
LLMs are required to have the human touch, empathy, and comprehension that human counselors can provide to enable more effective counseling. 
Reinforcement learning is adopted to facilitate empathic conversations~\citep{sharma2021towards}, generate motivational and empathetic responses with long-term reward maximization~\citep{saha2022towards}, and promote polite and empathetic counseling~\citep{mishra2023help}. 
Reinforcement learning in combination with LLMs can enhance the potential for a better dialogue system and reinforce counseling strategies in mental health. 
\citet{ji-towards-intention-understanding} showcased that language models struggle to comprehend user intentions and can inadvertently generate harmful or hateful content. 
In such cases, it becomes essential to employ contextual intent understanding, model the intention awareness~\citep{cambria2023seven}, and reason to identify the root causes of mental conditions, which could be used to enable empathetic conversational chatbot~\citep{li2021towards}, and generate responses with human-consistent empathetic intents~\citep{chen2022emphi}.
These strategies highlight the importance of understanding why users turn to LLM-based counseling and what they anticipate, offering insights for the design and deployment of these systems and making them more humane and responsive.

\paragraph{Promises and Caveats}

The use of LLMs in mental health counseling brings both potential benefits and perils. 
Chatbots engage in complex conversations with mental health consumers but struggle to identify and respond effectively to signs of distress, and consumers react negatively to unhelpful or risky chatbot responses~\citep{de2022chatbots}.
The responses of some publicly available LLM-based chatbots, when presented with prompts of increasing depression severity and suicidality, failed to recognize the risk progression appropriately~\citep{heston2023evaluating}. 
Generative LLMs such as ChatGPT struggled to detect unsafe responses in mental health support dialogues~\citep{qiu2023benchmark}. 
\citet{cabrera2023ethical} examined ethical issues using chatbots for mental health, identifying 24 moral dilemmas that cut across bioethical principles.
LLMs as chatbots require regulation, but the unreliability stops them from applying to the real world~\citep{gilbert2023large}.
ChatCounselor~\citep{liu2023chatcounselor} conducted supervised fine-tuning of base LLMs on real-world conversations between consulting clients and professional psychologists, although only evaluated the performance with GPT-4 without human grounds. 
Deploying LLM-based technology at scale for mental health may pose risks of misuse and require careful development and ongoing evaluation more systematically. 
\citep{li2023understanding} studied an annotation framework for understanding counselors' strategies and client reactions. 
Interacting with LLMs can provide insights into tailoring the use of these models in mental health counseling and make LLMs more professional virtual counselors, leading to the need for interactive language processing~\citep{wang2023interactive}.
Human preference data tailored for mental health scenarios can also be used to train reinforcement learning models to enable helpful and harmless dialogue agents~\citep{bai2022training}.
LLM-based methods facilitate psychological intervention and educational outreach for non-professionals~\citep{fu2023enhancing} and meanwhile posit some potential risks such as unreliable generation~\citep{gilbert2023large} and weakness in assessment of risk progression~\citep{heston2023evaluating}.
To ensure safe usage, more rigorous improvements and tests are needed.

\section{Conclusion}
\label{sec:conclusion}

AI indeed has the potential to be a valuable tool for identifying and supporting individuals who may be facing mental health challenges on social media platforms. However, it is essential to acknowledge and address the current challenges and concerns associated with its use.
This paper discusses the problems associated with large language models in mental health applications and emphasizes the importance of conducting further research to enhance the safety and reliability of LLM-based methods. 
While this study raises concerns regarding their effectiveness and explainability, it is essential to clarify that its intention is not to discredit the existing efforts made in this field. 
Instead, the research works discussed in this paper are regarded as essential steps toward exploring LLMs' real-world applications.
Our perspectives on rethinking LLMs in mental health applications aim to encourage the research community to reflect more deeply on LLMs' applicability, accountability, trustworthiness, and reliability~\citep{cambria2023seven}.

Firstly, it is noteworthy to mention that the application of LLMs for generative prediction in mental health has made significant progress, albeit without achieving a breakthrough. Nevertheless, several crucial issues, such as the instability in generated predictions and the performance of the generation-as-prediction paradigm, continue to persist and remain unresolved.

Secondly, LLMs possess the capability to generate explanations during generative predictions. 
This feature provides supplementary information to support the predictions made by LLMs. 
However, it is important to note that this does not necessarily imply that the model is inherently interpretable when applied to mental health analysis.

Thirdly, LLMs have shown considerable promise in generating coherent and fluent textual content. 
This quality makes them viable candidates for automatic mental health counseling. 
Nonetheless, it is important to emphasize the need for further research and development to ensure that the generated content is genuinely helpful and harmless for safe and effective application in mental health scenarios.

In conclusion, LLMs represent a promising frontier in mental health, but they must be approached with caution and respect for ethical principles. 
Their role should be one of support for human experts, emphasizing their unique abilities in the field. 
Our perspectives serve an important but not exhaustive view of applying LLMs in mental health.
Notably, there are other important aspects and considerations, such as cultural sensitivity, privacy, and data security. 
Robustness, ethical guidelines, and careful monitoring are essential components of deploying LLMs in the crucial task of addressing challenges in computational methods for mental health.

\section*{Ethical Considerations}
Ethical considerations are undeniably pivotal in deploying LLMs in mental health applications. Guaranteeing user safety, preserving privacy, and mitigating biases in responses are critical aspects that demand careful attention.
This paper underscores the exclusive reliance on publicly available publications for its research. 
Furthermore, it is essential to emphasize that there is no engagement in efforts to identify or directly interact with the individuals behind the social media posts. 
For research involving LLMs engaging in interactions with human beings, it is important to adhere to the user guidelines provided by the corresponding LLMs and to uphold the principles of research ethics rigorously. 
This ensures that ethical standards are maintained in all aspects of LLM-based research, especially in sensitive contexts like mental health.

\bibliography{mental-healthcare}

\begin{thebibliography}{50}
\providecommand{\natexlab}[1]{#1}
\providecommand{\url}[1]{\texttt{#1}}
\expandafter\ifx\csname urlstyle\endcsname\relax
  \providecommand{\doi}[1]{doi: #1}\else
  \providecommand{\doi}{doi: \begingroup \urlstyle{rm}\Url}\fi

\bibitem[Althoff et~al.(2016)Althoff, Clark, and Leskovec]{althoff2016large}
T.~Althoff, K.~Clark, and J.~Leskovec.
\newblock Large-scale analysis of counseling conversations: An application of
  natural language processing to mental health.
\newblock \emph{Transactions of the Association for Computational Linguistics},
  4:\penalty0 463--476, 2016.

\bibitem[Amin et~al.(2023)Amin, Cambria, and Schuller]{amin2023will}
M.~M. Amin, E.~Cambria, and B.~W. Schuller.
\newblock Will affective computing emerge from foundation models and general
  artificial intelligence? a first evaluation of {ChatGPT}.
\newblock \emph{IEEE Intelligent Systems}, 38\penalty0 (2):\penalty0 15--23,
  2023.

\bibitem[Bai et~al.(2022)Bai, Jones, Ndousse, Askell, Chen, DasSarma, Drain,
  Fort, Ganguli, Henighan, Joseph, Kadavath, Kernion, Conerly, El-Showk,
  Elhage, Hatfield-Dodds, Hernandez, Hume, Johnston, Kravec, Lovitt, Nanda,
  Olsson, Amodei, Brown, Clark, McCandlish, Olah, Mann, and
  Kaplan]{bai2022training}
Y.~Bai, A.~Jones, K.~Ndousse, A.~Askell, A.~Chen, N.~DasSarma, D.~Drain,
  S.~Fort, D.~Ganguli, T.~Henighan, N.~Joseph, S.~Kadavath, J.~Kernion,
  T.~Conerly, S.~El-Showk, N.~Elhage, Z.~Hatfield-Dodds, D.~Hernandez, T.~Hume,
  S.~Johnston, S.~Kravec, L.~Lovitt, N.~Nanda, C.~Olsson, D.~Amodei, T.~Brown,
  J.~Clark, S.~McCandlish, C.~Olah, B.~Mann, and J.~Kaplan.
\newblock Training a helpful and harmless assistant with reinforcement learning
  from human feedback.
\newblock \emph{arXiv preprint arXiv:2204.05862}, 2022.

\bibitem[Bhaumik et~al.(2023)Bhaumik, Srivastava, Jalali, Ghosh, and
  Chandrasekharan]{bhaumik2023mindwatch}
R.~Bhaumik, V.~Srivastava, A.~Jalali, S.~Ghosh, and R.~Chandrasekharan.
\newblock {MindWatch}: A smart cloud-based {AI} solution for suicide ideation
  detection leveraging large language models.
\newblock \emph{medRxiv}, 2023.
\newblock \doi{10.1101/2023.09.25.23296062}.
\newblock URL
  \url{https://www.medrxiv.org/content/early/2023/09/26/2023.09.25.23296062}.

\bibitem[Bibal et~al.(2022)Bibal, Cardon, Alfter, Wilkens, Wang,
  Fran{\c{c}}ois, and Watrin]{bibal-etal-2022-attention}
A.~Bibal, R.~Cardon, D.~Alfter, R.~Wilkens, X.~Wang, T.~Fran{\c{c}}ois, and
  P.~Watrin.
\newblock Is attention explanation? an introduction to the debate.
\newblock In \emph{Proceedings of the 60th Annual Meeting of the Association
  for Computational Linguistics (Volume 1: Long Papers)}, pages 3889--3900,
  Dublin, Ireland, May 2022. Association for Computational Linguistics.
\newblock \doi{10.18653/v1/2022.acl-long.269}.
\newblock URL \url{https://aclanthology.org/2022.acl-long.269}.

\bibitem[Brown et~al.(2020)Brown, Mann, Ryder, Subbiah, Kaplan, Dhariwal,
  Neelakantan, Shyam, Sastry, Askell, et~al.]{brown2020language}
T.~Brown, B.~Mann, N.~Ryder, M.~Subbiah, J.~D. Kaplan, P.~Dhariwal,
  A.~Neelakantan, P.~Shyam, G.~Sastry, A.~Askell, et~al.
\newblock Language models are few-shot learners.
\newblock \emph{Advances in Neural Information Processing Systems},
  33:\penalty0 1877--1901, 2020.

\bibitem[Cabrera et~al.(2023)Cabrera, Loyola, Maga{\~n}a, and
  Rojas]{cabrera2023ethical}
J.~Cabrera, M.~S. Loyola, I.~Maga{\~n}a, and R.~Rojas.
\newblock Ethical dilemmas, mental health, artificial intelligence, and
  {LLM}-based chatbots.
\newblock In \emph{International Work-Conference on Bioinformatics and
  Biomedical Engineering}, pages 313--326. Springer, 2023.

\bibitem[Cambria et~al.(2023)Cambria, Mao, Chen, Wang, and
  Ho]{cambria2023seven}
E.~Cambria, R.~Mao, M.~Chen, Z.~Wang, and S.-B. Ho.
\newblock Seven pillars for the future of artificial intelligence.
\newblock \emph{IEEE Intelligent Systems}, 38\penalty0 (6), 2023.

\bibitem[Chen et~al.(2022)Chen, Li, and Yang]{chen2022emphi}
M.~Y. Chen, S.~Li, and Y.~Yang.
\newblock {EmpHi}: Generating empathetic responses with human-like intents.
\newblock In \emph{Proceedings of the 2022 Conference of the North American
  Chapter of the Association for Computational Linguistics: Human Language
  Technologies}, pages 1063--1074, 2022.

\bibitem[Dai et~al.(2023)Dai, Sun, Dong, Hao, Ma, Sui, and Wei]{dai2023can}
D.~Dai, Y.~Sun, L.~Dong, Y.~Hao, S.~Ma, Z.~Sui, and F.~Wei.
\newblock Why can {GPT} learn in-context? language models secretly perform
  gradient descent as meta-optimizers.
\newblock In A.~Rogers, J.~Boyd-Graber, and N.~Okazaki, editors, \emph{Findings
  of the Association for Computational Linguistics: ACL 2023}, pages
  4005--4019, Toronto, Canada, July 2023. Association for Computational
  Linguistics.
\newblock \doi{10.18653/v1/2023.findings-acl.247}.
\newblock URL \url{https://aclanthology.org/2023.findings-acl.247}.

\bibitem[De~Freitas et~al.(2022)De~Freitas, U{\u{g}}uralp,
  O{\u{g}}uz-U{\u{g}}uralp, and Puntoni]{de2022chatbots}
J.~De~Freitas, A.~K. U{\u{g}}uralp, Z.~O{\u{g}}uz-U{\u{g}}uralp, and
  S.~Puntoni.
\newblock Chatbots and mental health: Insights into the safety of generative
  {AI}.
\newblock \emph{Journal of Consumer Psychology}, 2022.

\bibitem[Fitzpatrick et~al.(2017)Fitzpatrick, Darcy, and
  Vierhile]{fitzpatrick2017delivering}
K.~K. Fitzpatrick, A.~Darcy, and M.~Vierhile.
\newblock Delivering cognitive behavior therapy to young adults with symptoms
  of depression and anxiety using a fully automated conversational agent
  ({Woebot}): a randomized controlled trial.
\newblock \emph{JMIR mental health}, 4\penalty0 (2):\penalty0 e7785, 2017.

\bibitem[Fu et~al.(2023)Fu, Zhao, Li, Luo, Song, Zhai, Liu, Wang, Wang, Cheng,
  Zhang, and Yang]{fu2023enhancing}
G.~Fu, Q.~Zhao, J.~Li, D.~Luo, C.~Song, W.~Zhai, S.~Liu, F.~Wang, Y.~Wang,
  L.~Cheng, J.~Zhang, and B.~X. Yang.
\newblock Enhancing psychological counseling with large language model: A
  multifaceted decision-support system for non-professionals, 2023.

\bibitem[Gilbert et~al.(2023)Gilbert, Harvey, Melvin, Vollebregt, and
  Wicks]{gilbert2023large}
S.~Gilbert, H.~Harvey, T.~Melvin, E.~Vollebregt, and P.~Wicks.
\newblock Large language model {AI} chatbots require approval as medical
  devices.
\newblock \emph{Nature Medicine}, pages 1--3, 2023.

\bibitem[Harrigian et~al.(2021)Harrigian, Aguirre, and
  Dredze]{harrigian2021state}
K.~Harrigian, C.~Aguirre, and M.~Dredze.
\newblock On the state of social media data for mental health research.
\newblock In \emph{Proceedings of the Seventh Workshop on Computational
  Linguistics and Clinical Psychology: Improving Access}, pages 15--24. ACL,
  2021.

\bibitem[Heston(2023)]{heston2023evaluating}
T.~F. Heston.
\newblock Evaluating risk progression in mental health chatbots using
  escalating prompts.
\newblock \emph{medRxiv}, 2023.
\newblock \doi{10.1101/2023.09.10.23295321}.
\newblock URL
  \url{https://www.medrxiv.org/content/early/2023/09/12/2023.09.10.23295321}.

\bibitem[Ji(2022)]{ji-towards-intention-understanding}
S.~Ji.
\newblock Towards intention understanding in suicidal risk assessment with
  natural language processing.
\newblock In \emph{Findings of EMNLP}, pages 4028--4038. Association for
  Computational Linguistics, 2022.
\newblock URL \url{https://aclanthology.org/2022.findings-emnlp.297}.

\bibitem[Ji et~al.(2022)Ji, Zhang, Ansari, Fu, Tiwari, and
  Cambria]{ji2022mentalbert}
S.~Ji, T.~Zhang, L.~Ansari, J.~Fu, P.~Tiwari, and E.~Cambria.
\newblock {MentalBERT: Publicly Available Pretrained Language Models for Mental
  Healthcare}.
\newblock In \emph{Proceedings of LREC}, pages 7184--7190, Marseille, France,
  2022. European Language Resources Association.
\newblock URL \url{https://aclanthology.org/2022.lrec-1.778}.

\bibitem[Ji et~al.(2023)Ji, Zhang, Yang, Ananiadou, Cambria, and
  Tiedemann]{ji-domain-specific-longformer}
S.~Ji, T.~Zhang, K.~Yang, S.~Ananiadou, E.~Cambria, and J.~Tiedemann.
\newblock Domain-specific continued pretraining of language models for
  capturing long context in mental health.
\newblock \emph{arXiv preprint arXiv:2304.10447}, 2023.
\newblock URL \url{https://arxiv.org/abs/2304.10447}.

\bibitem[Joyce et~al.(2023)Joyce, Kormilitzin, Smith, and
  Cipriani]{joyce2023explainable}
D.~W. Joyce, A.~Kormilitzin, K.~A. Smith, and A.~Cipriani.
\newblock Explainable artificial intelligence for mental health through
  transparency and interpretability for understandability.
\newblock \emph{npj Digital Medicine}, 6\penalty0 (1):\penalty0 6, 2023.

\bibitem[Kroenke et~al.(2001)Kroenke, Spitzer, and Williams]{kroenke2001phq}
K.~Kroenke, R.~L. Spitzer, and J.~B. Williams.
\newblock The {PHQ}-9: validity of a brief depression severity measure.
\newblock \emph{Journal of General Internal Medicine}, 16\penalty0
  (9):\penalty0 606--613, 2001.

\bibitem[Lai et~al.(2023)Lai, Shi, Du, Wu, Fu, Dou, and Wang]{lai2023psyllm}
T.~Lai, Y.~Shi, Z.~Du, J.~Wu, K.~Fu, Y.~Dou, and Z.~Wang.
\newblock {Psy-LLM}: Scaling up global mental health psychological services
  with {AI}-based large language models, 2023.

\bibitem[Lan et~al.(2019)Lan, Liu, Zhou, and Yosinski]{lan2019lca}
J.~Lan, R.~Liu, H.~Zhou, and J.~Yosinski.
\newblock {LCA}: Loss change allocation for neural network training.
\newblock \emph{Advances in Neural Information Processing Systems}, 32, 2019.

\bibitem[Li et~al.(2023)Li, Ma, Mei, He, Zhang, Qiu, and
  Lan]{li2023understanding}
A.~Li, L.~Ma, Y.~Mei, H.~He, S.~Zhang, H.~Qiu, and Z.~Lan.
\newblock Understanding client reactions in online mental health counseling.
\newblock In \emph{Proceedings of the 61st Annual Meeting of the Association
  for Computational Linguistics (Volume 1: Long Papers)}, pages 10358--10376,
  2023.

\bibitem[Li et~al.(2021)Li, Li, Ning, Xia, Guo, Wei, Cui, and
  Wang]{li2021towards}
Y.~Li, K.~Li, H.~Ning, X.~Xia, Y.~Guo, C.~Wei, J.~Cui, and B.~Wang.
\newblock Towards an online empathetic chatbot with emotion causes.
\newblock In \emph{Proceedings of the 44th International ACM SIGIR Conference
  on Research and Development in Information Retrieval}, pages 2041--2045,
  2021.

\bibitem[Liu et~al.(2023{\natexlab{a}})Liu, Li, Cao, Ren, Liao, and
  Wu]{liu2023chatcounselor}
J.~M. Liu, D.~Li, H.~Cao, T.~Ren, Z.~Liao, and J.~Wu.
\newblock {ChatCounselor}: A large language models for mental health support.
\newblock \emph{arXiv preprint arXiv:2309.15461}, 2023{\natexlab{a}}.

\bibitem[Liu et~al.(2023{\natexlab{b}})Liu, Lin, Hewitt, Paranjape, Bevilacqua,
  Petroni, and Liang]{liu2023lost}
N.~F. Liu, K.~Lin, J.~Hewitt, A.~Paranjape, M.~Bevilacqua, F.~Petroni, and
  P.~Liang.
\newblock Lost in the middle: How language models use long contexts.
\newblock \emph{arXiv preprint arXiv:2307.03172}, 2023{\natexlab{b}}.

\bibitem[Min et~al.(2022)Min, Lyu, Holtzman, Artetxe, Lewis, Hajishirzi, and
  Zettlemoyer]{min-etal-2022-rethinking}
S.~Min, X.~Lyu, A.~Holtzman, M.~Artetxe, M.~Lewis, H.~Hajishirzi, and
  L.~Zettlemoyer.
\newblock Rethinking the role of demonstrations: What makes in-context learning
  work?
\newblock In Y.~Goldberg, Z.~Kozareva, and Y.~Zhang, editors, \emph{Proceedings
  of the 2022 Conference on Empirical Methods in Natural Language Processing},
  pages 11048--11064, Abu Dhabi, United Arab Emirates, Dec. 2022. Association
  for Computational Linguistics.
\newblock \doi{10.18653/v1/2022.emnlp-main.759}.
\newblock URL \url{https://aclanthology.org/2022.emnlp-main.759}.

\bibitem[Mishra et~al.(2023)Mishra, Priya, and Ekbal]{mishra2023help}
K.~Mishra, P.~Priya, and A.~Ekbal.
\newblock Help me heal: A reinforced polite and empathetic mental health and
  legal counseling dialogue system for crime victims.
\newblock \emph{Proceedings of the AAAI Conference on Artificial Intelligence},
  37\penalty0 (12):\penalty0 14408--14416, Jun. 2023.
\newblock \doi{10.1609/aaai.v37i12.26685}.
\newblock URL \url{https://ojs.aaai.org/index.php/AAAI/article/view/26685}.

\bibitem[M{\"o}kander et~al.(2023)M{\"o}kander, Schuett, Kirk, and
  Floridi]{mokander2023auditing}
J.~M{\"o}kander, J.~Schuett, H.~R. Kirk, and L.~Floridi.
\newblock Auditing large language models: a three-layered approach.
\newblock \emph{AI and Ethics}, pages 1--31, 2023.

\bibitem[Naseem et~al.(2022)Naseem, Lee, Khushi, Kim, and
  Dunn]{naseem-etal-2022-benchmarking}
U.~Naseem, B.~C. Lee, M.~Khushi, J.~Kim, and A.~Dunn.
\newblock Benchmarking for public health surveillance tasks on social media
  with a domain-specific pretrained language model.
\newblock In \emph{Proceedings of NLP Power! The First Workshop on Efficient
  Benchmarking in NLP}, pages 22--31, Dublin, Ireland, May 2022. Association
  for Computational Linguistics.
\newblock \doi{10.18653/v1/2022.nlppower-1.3}.
\newblock URL \url{https://aclanthology.org/2022.nlppower-1.3}.

\bibitem[Pavalanathan and De~Choudhury(2015)]{pavalanathan2015identity}
U.~Pavalanathan and M.~De~Choudhury.
\newblock Identity management and mental health discourse in social media.
\newblock In \emph{WWW}, pages 315--321. ACM, 2015.

\bibitem[Qiu et~al.(2023{\natexlab{a}})Qiu, He, Zhang, Li, and
  Lan]{qiu2023smile}
H.~Qiu, H.~He, S.~Zhang, A.~Li, and Z.~Lan.
\newblock {SMILE}: Single-turn to multi-turn inclusive language expansion via
  {ChatGPT} for mental health support.
\newblock \emph{arXiv preprint arXiv:2305.00450}, 2023{\natexlab{a}}.

\bibitem[Qiu et~al.(2023{\natexlab{b}})Qiu, Zhao, Li, Zhang, He, and
  Lan]{qiu2023benchmark}
H.~Qiu, T.~Zhao, A.~Li, S.~Zhang, H.~He, and Z.~Lan.
\newblock A benchmark for understanding dialogue safety in mental health
  support.
\newblock In \emph{CCF International Conference on Natural Language Processing
  and Chinese Computing}, pages 1--13. Springer, 2023{\natexlab{b}}.

\bibitem[Rudin(2019)]{rudin2019stop}
C.~Rudin.
\newblock Stop explaining black box machine learning models for high stakes
  decisions and use interpretable models instead.
\newblock \emph{Nature Machine Intelligence}, 1\penalty0 (5):\penalty0
  206--215, 2019.

\bibitem[Saha et~al.(2022)Saha, Gakhreja, Das, Chakraborty, and
  Saha]{saha2022towards}
T.~Saha, V.~Gakhreja, A.~S. Das, S.~Chakraborty, and S.~Saha.
\newblock Towards motivational and empathetic response generation in online
  mental health support.
\newblock In \emph{Proceedings of the 45th international ACM SIGIR conference
  on research and development in information retrieval}, pages 2650--2656,
  2022.

\bibitem[Sarkar et~al.(2023)Sarkar, Gaur, Chen, Garg, and
  Srivastava]{sarkar2023review}
S.~Sarkar, M.~Gaur, L.~K. Chen, M.~Garg, and B.~Srivastava.
\newblock A review of the explainability and safety of conversational agents
  for mental health to identify avenues for improvement.
\newblock \emph{Frontiers in Artificial Intelligence}, 6, 2023.

\bibitem[Sharma et~al.(2021)Sharma, Lin, Miner, Atkins, and
  Althoff]{sharma2021towards}
A.~Sharma, I.~W. Lin, A.~S. Miner, D.~C. Atkins, and T.~Althoff.
\newblock Towards facilitating empathic conversations in online mental health
  support: A reinforcement learning approach.
\newblock In \emph{Proceedings of the Web Conference}, pages 194--205, 2021.

\bibitem[Shi et~al.(2023)Shi, Chen, Misra, Scales, Dohan, Chi, Sch{\"a}rli, and
  Zhou]{shi2023large}
F.~Shi, X.~Chen, K.~Misra, N.~Scales, D.~Dohan, E.~H. Chi, N.~Sch{\"a}rli, and
  D.~Zhou.
\newblock Large language models can be easily distracted by irrelevant context.
\newblock In \emph{International Conference on Machine Learning}, pages
  31210--31227. PMLR, 2023.

\bibitem[Turpin et~al.(2023)Turpin, Michael, Perez, and
  Bowman]{turpin2023language}
M.~Turpin, J.~Michael, E.~Perez, and S.~R. Bowman.
\newblock Language models don't always say what they think: Unfaithful
  explanations in chain-of-thought prompting.
\newblock \emph{arXiv preprint arXiv:2305.04388}, 2023.

\bibitem[Vajre et~al.(2021)Vajre, Naylor, Kamath, and
  Shehu]{vajre2021psychbert}
V.~Vajre, M.~Naylor, U.~Kamath, and A.~Shehu.
\newblock {PsychBERT}: a mental health language model for social media mental
  health behavioral analysis.
\newblock In \emph{2021 IEEE International Conference on Bioinformatics and
  Biomedicine (BIBM)}, pages 1077--1082. IEEE, 2021.

\bibitem[Von~Oswald et~al.(2023)Von~Oswald, Niklasson, Randazzo, Sacramento,
  Mordvintsev, Zhmoginov, and Vladymyrov]{von2023transformers}
J.~Von~Oswald, E.~Niklasson, E.~Randazzo, J.~Sacramento, A.~Mordvintsev,
  A.~Zhmoginov, and M.~Vladymyrov.
\newblock Transformers learn in-context by gradient descent.
\newblock In \emph{International Conference on Machine Learning}, pages
  35151--35174. PMLR, 2023.

\bibitem[Wang et~al.(2023)Wang, Zhang, Yang, Shi, Zhou, Hao, Xiong, Li, Sim,
  Chen, Zhu, Yang, Nik, Liu, Lin, Wang, Liu, Chen, Xu, Liu, Guo, and
  Fu]{wang2023interactive}
Z.~Wang, G.~Zhang, K.~Yang, N.~Shi, W.~Zhou, S.~Hao, G.~Xiong, Y.~Li, M.~Y.
  Sim, X.~Chen, Q.~Zhu, Z.~Yang, A.~Nik, Q.~Liu, C.~Lin, S.~Wang, R.~Liu,
  W.~Chen, K.~Xu, D.~Liu, Y.~Guo, and J.~Fu.
\newblock Interactive natural language processing.
\newblock \emph{arXiv preprint arXiv:2305.13246}, 2023.

\bibitem[Wei et~al.(2022)Wei, Wang, Schuurmans, Bosma, Xia, Chi, Le, Zhou,
  et~al.]{wei2022chain}
J.~Wei, X.~Wang, D.~Schuurmans, M.~Bosma, F.~Xia, E.~Chi, Q.~V. Le, D.~Zhou,
  et~al.
\newblock Chain-of-thought prompting elicits reasoning in large language
  models.
\newblock \emph{Advances in Neural Information Processing Systems},
  35:\penalty0 24824--24837, 2022.

\bibitem[Wu and Varshney(2023)]{wu2023meta}
X.~Wu and L.~R. Varshney.
\newblock A meta-learning perspective on transformers for causal language
  modeling.
\newblock \emph{arXiv preprint arXiv:2310.05884}, 2023.

\bibitem[Xu et~al.(2023)Xu, Yao, Dong, Gabriel, Yu, Hendler, Ghassemi, Dey, and
  Wang]{xu2023mental}
X.~Xu, B.~Yao, Y.~Dong, S.~Gabriel, H.~Yu, J.~Hendler, M.~Ghassemi, A.~K. Dey,
  and D.~Wang.
\newblock {Mental-LLM}: Leveraging large language models for mental health
  prediction via online text data.
\newblock \emph{arXiv preprint arXiv:2307.14385}, 2023.

\bibitem[Yang et~al.(2023{\natexlab{a}})Yang, Ji, Zhang, Xie, Kuang, and
  Ananiadou]{yang2023interpretable}
K.~Yang, S.~Ji, T.~Zhang, Q.~Xie, Z.~Kuang, and S.~Ananiadou.
\newblock Towards interpretable mental health analysis with large language
  models.
\newblock In \emph{Proceedings of EMNLP}, 2023{\natexlab{a}}.
\newblock URL \url{https://arxiv.org/abs/2304.03347}.

\bibitem[Yang et~al.(2023{\natexlab{b}})Yang, Zhang, Kuang, Xie, and
  Ananiadou]{yang2023mentalllama}
K.~Yang, T.~Zhang, Z.~Kuang, Q.~Xie, and S.~Ananiadou.
\newblock {MentalLLaMA}: Interpretable mental health analysis on social media
  with large language models.
\newblock \emph{arXiv preprint arXiv:2309.13567}, 2023{\natexlab{b}}.

\bibitem[Zhang et~al.(2022)Zhang, Schoene, Ji, and Ananiadou]{zhang2022natural}
T.~Zhang, A.~Schoene, S.~Ji, and S.~Ananiadou.
\newblock Natural language processing applied to mental illness detection: A
  narrative review.
\newblock \emph{npj Digital Medicine}, 5, 2022.

\bibitem[Zhao et~al.(2023)Zhao, Chen, Yang, Liu, Deng, Cai, Wang, Yin, and
  Du]{zhao2023explainability}
H.~Zhao, H.~Chen, F.~Yang, N.~Liu, H.~Deng, H.~Cai, S.~Wang, D.~Yin, and M.~Du.
\newblock Explainability for large language models: A survey.
\newblock \emph{arXiv preprint arXiv:2309.01029}, 2023.

\end{thebibliography}

\end{document}